\documentclass[sigconf]{acmart}

\usepackage{array}
\usepackage{float}
\usepackage{amsmath}
\usepackage[colorinlistoftodos]{todonotes}
\usepackage[ruled,linesnumbered]{algorithm2e}
\SetKwInOut{Input}{Given}
\SetKwInOut{Output}{Output}
\SetKwComment{Comment}{/* }{ */}
\usepackage{longtable}
\usepackage{enumitem}
\usepackage{makecell}
\usepackage{bbm}

\AtBeginDocument{%
  }

\setcopyright{acmcopyright}
\copyrightyear{2023}
\acmYear{2023}
\acmDOI{XXXXXXX.XXXXXXX}

\acmConference[Conference acronym 'XX]{2023}
\acmISBN{978-1-4503-XXXX-X/18/06}

\begin{document}

\title{XUAT-Copilot: Multi-Agent Collaborative System for Automated User Acceptance Testing with Large Language Model}

\author{Zhitao Wang$^*$, Wei Wang$^*$, Zirao Li, Long Wang, Can Yi, Xinjie Xu, Luyang Cao, Hanjing Su, Shouzhi Chen, Jun Zhou}
\email{{zhitaowang, dylanwwang, ziraoli, oliverlwang}@tencent.com}
\affiliation{%
  \institution{WeChat Pay, Tencent}
  \city{Shenzhen}
  \country{China}
}

\renewcommand{\shortauthors}{Zhitao Wang et al.}

\begin{abstract}
In past years, we have been dedicated to automating user acceptance testing (UAT) process of WeChat Pay, one of the most influential mobile payment applications in China. A system titled XUAT has been developed for this purpose. However, there is still a human-labor-intensive stage, i.e., test scripts generation, in the current system. Therefore, in this paper, we concentrate on methods of boosting the automation level of the current system, particularly the stage of test scripts generation. With recent notable successes, large language models (LLMs) demonstrate significant potential in attaining human-like intelligence and there has been a growing research area that employs LLMs as autonomous agents to obtain human-like decision-making capabilities. Inspired by these works, we propose an LLM-powered multi-agent collaborative system, named XUAT-Copilot, for automated UAT. The proposed system mainly consists of three LLM-based agents responsible for action planning, state checking and parameter selecting, respectively, and two additional modules for state sensing and case rewriting. The agents interact with testing device, make human-like decision and generate action command in a collaborative way. The proposed multi-agent system achieves a close effectiveness to human testers in our experimental studies and gains a significant improvement of $Pass@1$ accuracy compared with single-agent architecture. More importantly, the proposed system has launched in the formal testing environment of WeChat Pay mobile app, which saves a considerable amount of manpower in the daily development work. 
\end{abstract}



\keywords{Automated UAT, AI Agents, Large Language Model}


\maketitle
\def\thefootnote{*}\footnotetext{These authors contributed equally to this work}\def\thefootnote{\arabic{footnote}}
\section{Introduction}

User Acceptance Testing (UAT) is a critical phase in the software development life cycle that involves assessing the functionality, usability, and performance of a software application or system. The main goal of UAT is to evaluate whether the developed software meets the predefined requirements from the end-user's perspective. It validates that the software is fit for purpose and ready for deployment in a real-world environment. However, traditional UAT methodologies, which rely heavily on manual testing, are time-consuming, resource-intensive, and susceptible to human errors. It is thus non-trivial to explore the way of automating UAT to improve efficiency, accuracy, and overall software quality.

\begin{figure*}[htbp]
\centering
\includegraphics[width=18cm]{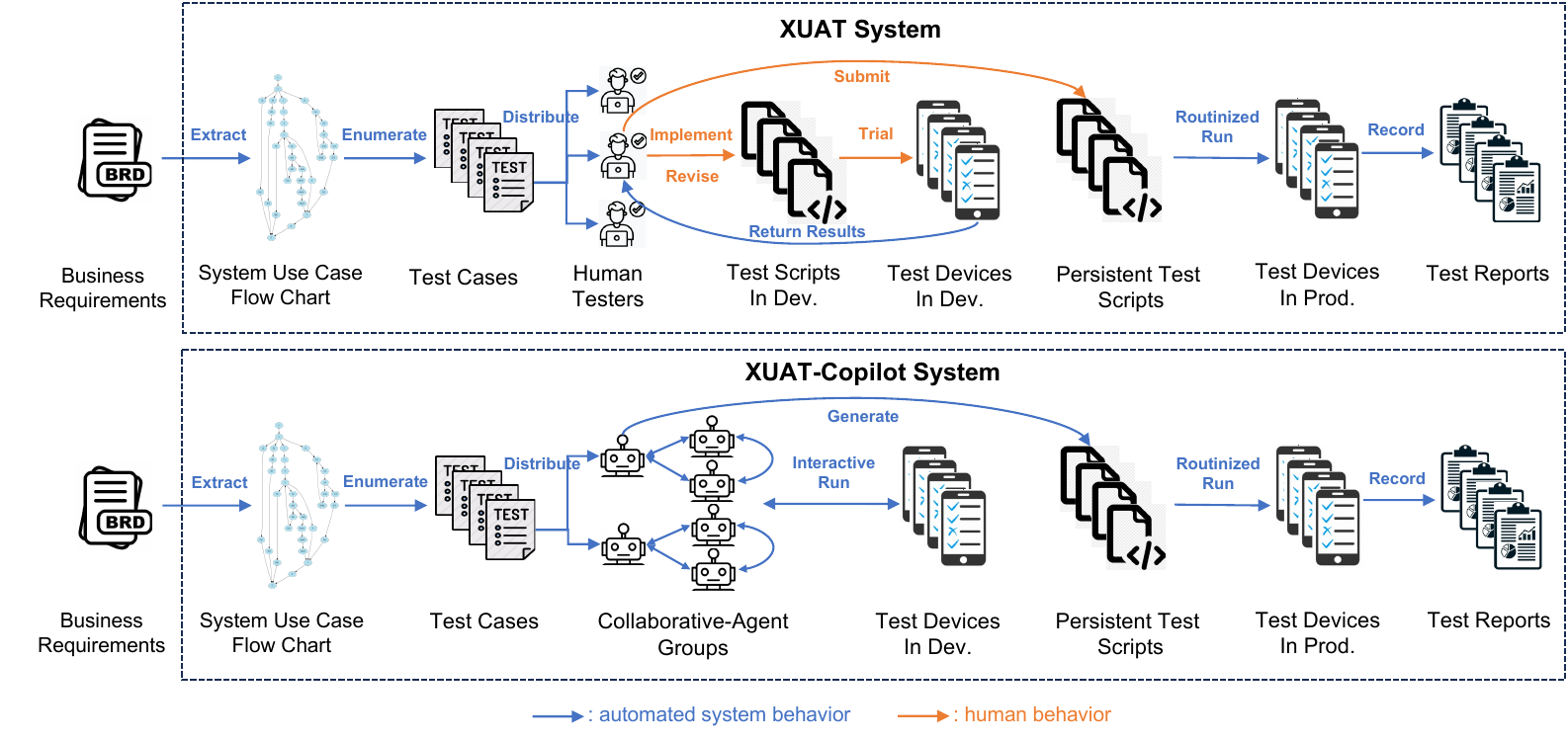}
\caption{The overview of XUAT system and XUAT-Copilot system}
\label{fig:sys}
\end{figure*}

WeChat Pay\footnote[1]{\url{https://pay.weixin.qq.com/index.php/public/wechatpay_en}}, as one of the most influential mobile payment applications in China, has become an integral part of daily life for billions of users. Ensuring the seamless functionality, usability, and compatibility of the application is of paramount importance for maintaining customer satisfaction and trust. Therefore, in past years, we have been dedicated to the development of an effective, comprehensive and low-cost UAT system, named XUAT. The workflow of the XUAT system is illustrated in Fig.\ref{fig:sys}. At the initial stage, business requirements are represented as use case flow charts, which are intended to ensure the completeness, accuracy and conformity of the requirements. Given the flow charts, the system enumerates all possible flows and transfers each flow as one specific test case in semi-structured natural language. At the following stage, these test cases are distributed to a number of human testers, who are responsible for implementing test scripts consisting of a series of Android Debug Bridge (ADB) commands to simulate end-users' actions expected in the test cases on the App (in this paper, we only focus on testing on Android system). In this process, human testers often make trial and errors on test devices to guarantee the script can be executed in expectation of the test case accurately. The final scripts submitted by testers (also called "persistent test scripts") will be automatically ran in routine by the system in the future and reports are recorded in the system. Based on the test reports, a group of acceptance engineers are asked to validate the results of test cases against the expectations of the requirements, and document any defects, issues or improvements.

Apparently, current version of XUAT is still a semi-automated system, which is particularly labor-intensive at the stage of test script implementation. 
Understanding contents of test cases and making trial and error are inevitable in this human-engaged process. This motivates us to seek for more automated and efficiency approaches of implementing test scripts.

Recent advancements in large language models (LLMs) like ChatGPT, have demonstrated remarkable capabilities in not just linguistic tasks, but also various general tasks especially embodied agents planning, where LLMs are applied for making goal-driven decisions in interactive and embodied environments \cite{song2022llm,singh2023progprompt,huang2022language,lin2023text2motion}. Test script generation in XUAT is also a task-oriented, interactive and embodied scenario, where end-users are expected to be represented by autonomous agents who automatically interact with the app through ADB commands to achieve the goal defined in the test case step-by-step. Inspired by the previous work, we are interested to investigate the potential of employing LLM powered agents to automate the test script generation process in the XUAT system. 

Despite of the power of LLMs, this work still faces a set of unique challenges:
\begin{itemize}[leftmargin=0.5cm]
\item \textbf{Highly condensed expression}. Each step of test cases is expressed in a highly brief way but often implies a series of actions. For example, the step of "Submitting Personal Information" expects a long action-sequence: clicking name input box, entering text of name, clicking ID input box, entering ID number, etc. 
\item \textbf{Context sensitive actions}. Similar or even same contents ask for different actions at different graphical user interface (GUI) pages or with different expectations. For example, the step of "Confirming the Information" indicates a single action of clicking the confirmation button on some feedback pages, but multiple actions of scrolling the screen, selecting the checkbox and clicking the confirmation button on some agreement pages. 
\item \textbf{Many test parameters}. In the WeChat Pay app, around 80 percent of actions belong to the type of "submitting information", which indicates hundreds of test parameters are maintained. In current system, a list of all test parameters is provided and human testers need to select appropriate parameters. This brings difficulties for LLM to recognize required parameters without experience of human testers.
\item \textbf{Step-by-step correctness}. Different from random test methods, UAT needs to strictly follow the provided test case such that the final goal can be achieved. If one step is wrong, the whole test case will not be passed. Therefore, it is crucial for agents to self-check step-by-step correctness of the test case.
\end{itemize}

\begin{figure*}[htbp]
\centering
\includegraphics[width=17cm]{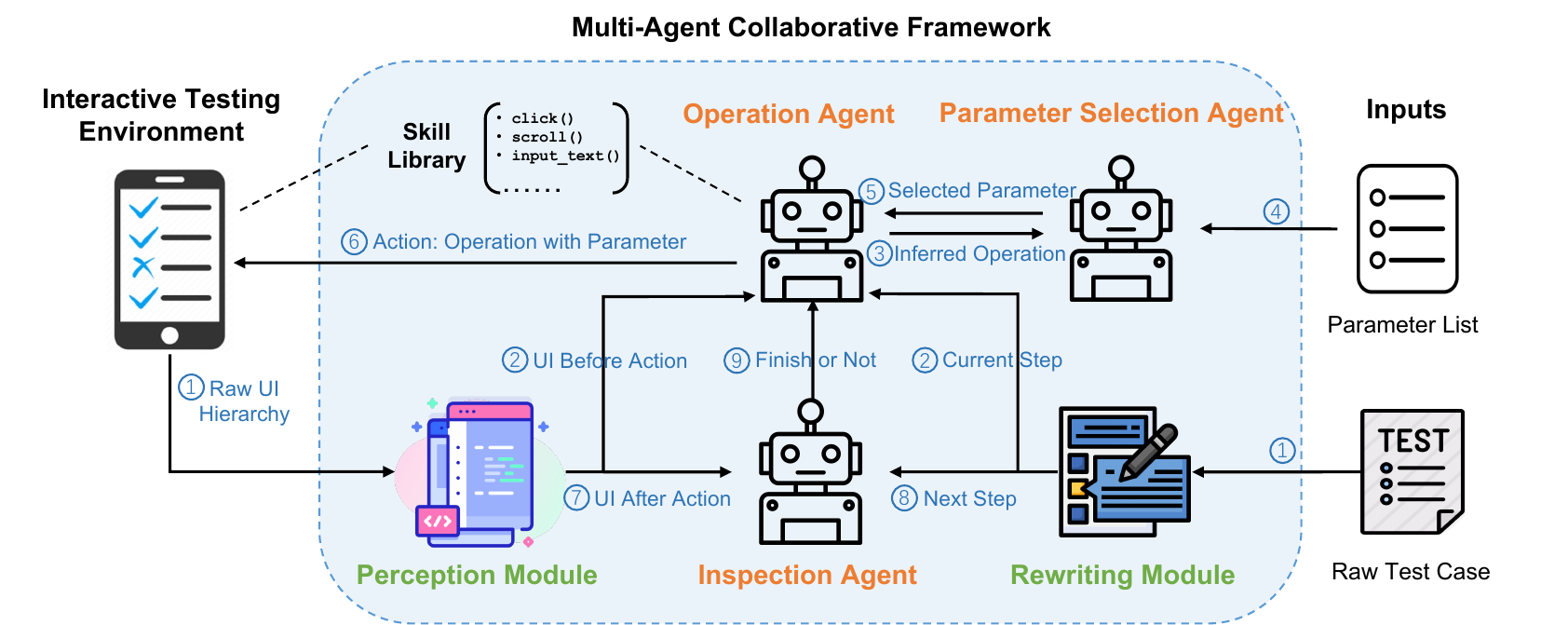}
\caption{The proposed multi-agent collaborative framework}
\label{fig:framework}
\end{figure*}

In this paper, we attempt to resolve above challenges and propose XUAT-Copilot, an LLM-powered multi-agent collaborative system for automating UAT. 
Fig \ref{fig:framework} illustrates the proposed multi-agent collaborative system, which mainly consists of three LLM-based agents responsible for action planning, state checking and parameter selecting, respectively, and two additional modules for state sensing and case rewriting. Specifically, \textbf{Rewriting Module} is applied to make the contents of test cases more understandable by the agents. On the other hand, \textbf{Perception Module} is utilized to provide precise and concise GUI information for the agents due to context sensitivity of this scenario. The three agents are associated in a collaborative way. Given the rewritten current step instruction of the test case, \textbf{Operation Agent} is responsible for planning expected user actions and generating corresponding action commands. If the commands require parameters, \textbf{Parameter Selection Agent} is asked by the operation agent for choosing appropriate parameters from hundreds of candidates. With returned parameters, the operation agent regenerates action commands and submits to execute them on testing devices simulating the expected user actions. Since each step may involve a series of user actions, \textbf{Inspection Agent} is devised to verify whether the state achieves the expectation of current step. Generally, if the APP navigates to the GUI page where actions of next step to be performed, it satisfies expectation of current step. Therefore, the inspection agent makes a decision based on both GUI information after actions and contents of next step. 

Based on the above collaborative multi-agent architecture, the proposed system achieves a close effectiveness to human testers in our experimental studies. Compared with single-agent architecture, our multi-agent architecture gains a significant improvement of Pass@1 and Complete@1 metrics. More importantly, the proposed XUAT-Copilot system has launched in the formal testing environment of WeChat Pay mobile app, which saves a considerable amount of manpower in the daily development work. 

The contributions of this paper could be concluded as follows:
\begin{itemize}[leftmargin=0.5cm]
\item To the best of our knowledge, we are the first to employ LLMs for UAT. We demonstrate the possibility and feasibility of LLMs to automate the test script generation in the UAT process.
\item We propose a novel multi-agent collaborative approach to UAT system that harnesses the power of pre-trained LLMs for both high-level human language instruction and low-level command planning and develop a range of innovative approaches aimed at enhancing the understanding and planning capabilities of LLMs in the context of the UAT scenario.
\item The proposed system has been applied to real research and development environment of WeChat Pay, a mobile payment app with a billion-level user base, with a significant improvement on testing efficiency.
\end{itemize}

\section{Background}
\subsection{Automated UAT and XUAT System}
UAT stands for User Acceptance Testing. It is a phase in the software development process where the end users or clients test the software to ensure it meets their requirements and expectations. The main goal of UAT is to validate that the software is functioning as intended and is ready for deployment in a real-world environment. Automated UAT aims to reduce the manual effort required to execute test cases, increase the efficiency and accuracy of the testing process, and shorten the overall testing cycle. It allows for faster feedback, easier identification of defects, and improved consistency in testing compared to manual UAT.

XUAT system is an instantiation of automated UAT on the WeChat Pay mobile app. As shown in Fig. \ref{fig:sys}, if business requirements are represented in predefined semi-structure, the XUAT system could automatically generate the test cases. However, the stage of transferring test cases to executable scripts requires intensive human participation. Human testers initially write the script, i.e., a series of ADB commands, and make a run trial on the development test devices to check whether the script simulates the expected actions and achieve the goal of the given test case. If there are some errors, testers need to make some revisions and a trial again, until return results satisfy the expectations. The final scripts submitted by testers are called as "persistent test scripts", since these scripts would be run in routine by the system and test reports are recorded automatically. In order to improve the system's automation level, in this paper, we focus on the most labor-intensive stage, i.e., test script implementation (generation), and apply large language models to make this stage automated. 

It should be noted that our work on automated UAT is very different from existing works on automated graphical user interface (GUI) testing \cite{li2019humanoid,pan2020reinforcement,liu2023fill,liu2023chatting}. First, the two testing processes have different purposes. Automated GUI testing is focused on testing the graphical user interface of an application, while automated UAT aims to validate the overall functionality and usability of the application from an end-user perspective. Second, the two testing processes have distinct paradigms. Automated GUI testing is often defined as a black-box task, thus methods on this task pay much attention to exploration of all UI events under a high degree of freedom. In contrast, automate UAT is a kind of white-box task, where testing behaviors are strictly constrained by well-predefined test cases such that the specific function points can be tested. Last, the two testing processes concentrate on different testing metrics. Previous works on automated GUI testing evaluate the performance by activity coverage, while UAT performance, in this paper, is evaluated by pass rates of test cases.

\subsection{Large Language Models}
Large Language Models (LLMs) are the latest breakthroughs in natural language processing. LLMs acquire the ability to achieve general-purpose language understanding and generation by training on massive amounts of data to learn billions of parameters. As autoregressive models, the training of LLMs is taking an input text and repeatedly predicting the next token or word. Most popular LLMs are based on the transformer architecture \cite{vaswani2017attention}.

At the beginning of the emergence of LLMs, fine tuning was the only way make a LLM could be adapt to specific tasks. Recent LLMs, such as GPT-3 \cite{brown2020language}, GPT-4 \cite{openai2023gpt4}, LLaMA \cite{touvron2023llama1,touvron2023llama2} and PaLM \cite{chowdhery2022palm,anil2023palm}, however, can achieve similar task-specific results without requiring additional training through prompt-engineering. Prompt engineering is the process of formulating a piece of text, which often describes the task and can be understood by LLMs. Prompt engineering is enabled by in-context learning \cite{dong2022survey}, defined as a model's ability to temporarily learn from prompts, which is an emergent ability of LLMs \cite{wei2022emergent}. Based on the remarkable general intelligence of LLMs, the application of in-context learning ability has shifted to zero-shot learning, where LLMs make predictions by directly describing the desired output. On the other hand, few-shot learning is utilized to augment the context with a few examples of desired inputs and outputs, which enables LLMs to recognize the input prompt syntax and patterns of the output. Instead of few-shot learning, which easily reaches to token limitation of LLMs, we apply zero-shot paradigm for the agents, which has shown satisfactory performance in the UAT scenario.

Chain-of-thought (CoT) \cite{wei2022chain} prompting is a technique that allows LLM to solve a problem as a series of intermediate steps. CoT prompting improves reasoning ability by inducing the model to answer a multi-step problem with steps of reasoning that mimic a train of thought. Several follow-up works have been extended to more sophisticated reasoning architecture beyond simple prompting. For example, Selection-Inference \cite{creswell2022selection} divides the reasoning process into two steps of “selection” and “inference”. Tree-of-Thought \cite{yao2023tree} generalizes over CoT prompting and encourages exploration over thoughts in a tree structure. In contrast, ReAct \cite{yao2022react} integrates model actions and corresponding observations into a coherent stream of inputs for the model to reason more accurately and tackle tasks beyond reasoning. UAT is also a kind of interactive decision making, therefore, in this paper, we employ a prompting schema similar to the ReACT with more complex structure using personas and a more comprehensive CoT prompting. Additionally, in order to reduce hallucinations and inefficient planning, we apply self-reflection mechanism \cite{shinn2023reflexion}, which prompts agents to self-reflect on task feedback and maintain reflective information in memory buffer to induce better decision-making in subsequent trials.

\subsection{Autonomous Agent and LLM-Based Agent}
An early definition of autonomous agent is a system situated within and a part of an environment that senses that environment and acts on it, over time, in pursuit of its own agenda and so as to effect what it senses in the future \cite{franklin1996agent}. Characteristics of ideal autonomous agents include: (a) Autonomy: They can operate independently, without human intervention, and make decisions based on their own internal logic; (b) Perception: They can perceive their environment, using sensors, cameras, or other input devices, and process the data; (c) Action: They can take actions in the physical or digital world, using actuators, motors, or other output devices, to achieve their goals; (d) Learning: They can learn from experience, adapt to changing circumstances, and improve their performance over time; (e) Communication: They can communicate with other agents, humans, or systems, using various protocols and interfaces. Therefore, invention of autonomous agents has long been recognized as a promising approach to achieving artificial general intelligence (AGI). 

In previous studies, autonomous agents are assumed to act based on policy functions, and learned in isolated and restricted environments \cite{mnih2015human,lillicrap2015continuous,schulman2017proximal,haarnoja2018soft}. The survey \cite{wang2023survey} has pointed out that such assumptions significantly differ from the human learning process, since individuals can learn from a much wider variety of environments. Due to these gaps, the agents learned from the previous studies are usually far from replicating human-level decision processes, especially in unconstrained, open-domain settings. As mentioned above, with notable successes, LLMs has demonstrated potential in achieving human-like intelligence and leads to a trending research area that employs LLMs to build autonomous agents to obtain human-like decision-making capabilities. In a very short time, researchers have developed numerous agents \cite{nakano2021webgpt,yao2022webshop,wang2023voyager,park2023generative,schick2023toolformer,shen2023hugginggpt,deng2023mind2web,wang2023recagent,qian2023communicative}. The key idea is equipping agents with crucial human capabilities like memory and planning by prompting, tuning or augmenting LLMs to complete various tasks. The initial inspiration of this paper comes from a group of LLM-based embodied agents \cite{song2022llm,singh2023progprompt,huang2022language,lin2023text2motion}, which are able to make goal-driven decisions and execute corresponding actions in interactive and embodied environments. 

A very recent series of research focus on developing LLM-powered multi-agent system. Most common interaction schema in these works is ordered cooperation \cite{chen2023autoagents,wu2023autogen,hong2023metagpt,talebirad2023multi}. Different agents participate the task in an ordered sequence, i.e., agent of current step receives the output from previous agent and generate its own output for next agent. Another schema is disordered cooperation, where agents cooperates with each other in a complex way instead of a simple pipeline and the cooperation structure may vary from to different tasks \cite{mandi2023roco,wang2023unleashing}. In contrast to cooperation, some works design an adversarial interaction schema, which put agents in adversarial relationship and push the agents to make advancements themselves \cite{chan2023chateval,du2023improving}. In this paper, we employ disordered cooperation schema and carefully devise the interaction structure for the proposed multi-agent system.

\begin{figure*}[htbp]
\centering
\includegraphics[width=17cm]{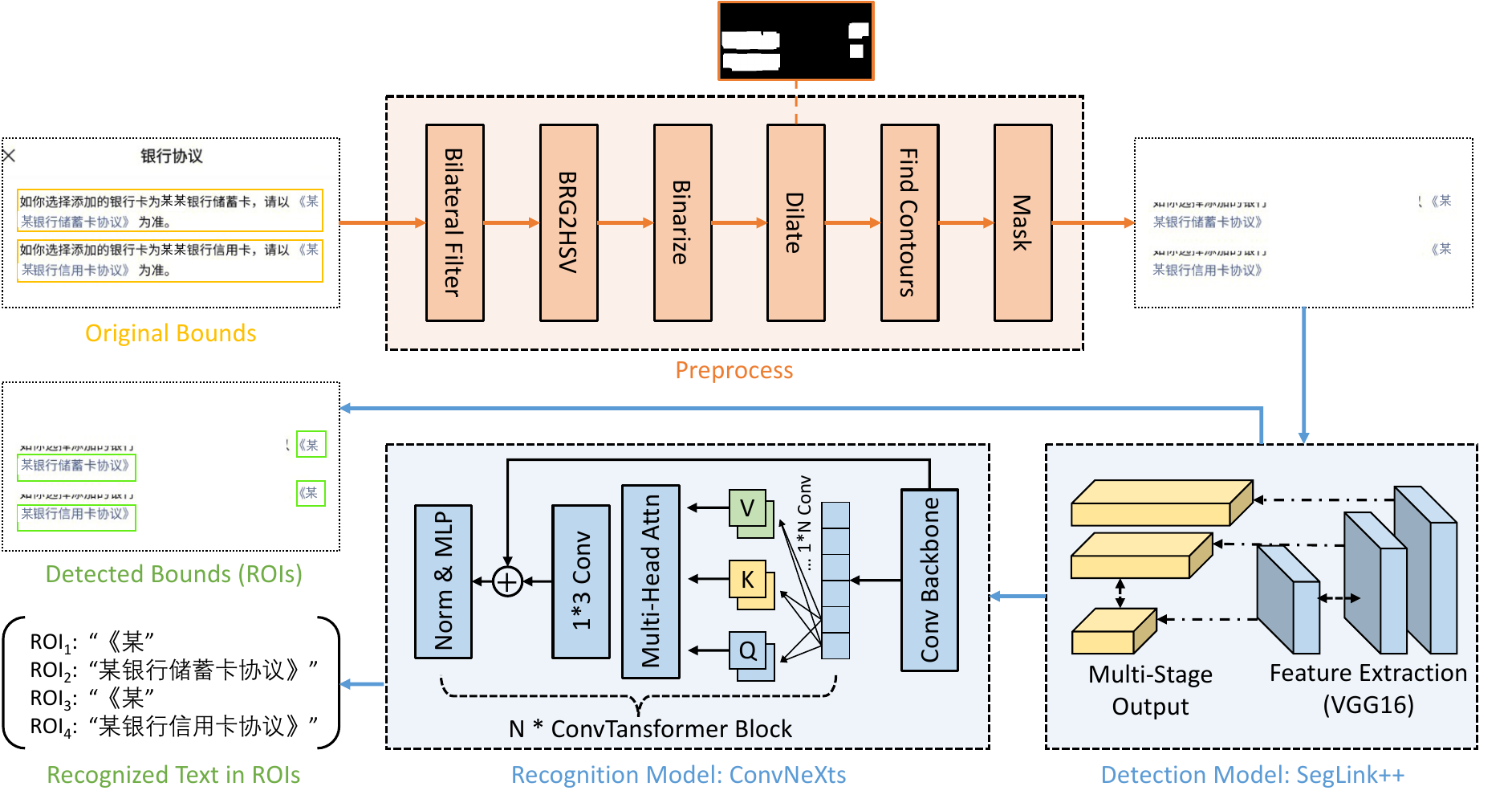}
\caption{Image Processing Pipeline For Widgets With Hyperlink}
\label{fig:hyperlink}
\end{figure*}

\section{XUAT-Copilot} 

\subsection{Problem Statement}
As shown in Fig. \ref{fig:framework}, the system receives two types of inputs and interacts with test environment through a predefined skill library. They are defined as follows: 
\begin{itemize}[leftmargin=0.5cm]
\item \textbf{Test Case} ($C$): Test case is an ordered sequence of instructions defined as $C = \left[c_1, ..., c_n \right]$, where $n$ is total steps of the test case $C$. Each piece of text instruction $c_i$ describes the expected goal implying user actions at $i$th step of the test case. 
\item \textbf{Parameter List} ($P$): Parameter list is associated with a specific test case denoted as  $P = \{(p_1, v_1), ..., (p_m, v_m)\}$, where $(p_i, v_i)$ denotes each possible test parameters and their corresponding values in the test case $S$ and $m$ is the number of all parameters. 
\item \textbf{Environment} ($E$): Environment in this scenario is a (virtual) mobile device, which is represented as $E$. The system needs to continuously interactive with the environment $E$ to execute actions as well as update the state of agents.
\item \textbf{Skill Library} ($F$): Skill library represents possible agents' interactive actions with the test environment, which are expected users' operations on the APP. We define a set of functions, which wrap ADB commands and simulate users' actions in the environment, as the skill library: $F = \{f_{click}, f_{scroll}, ...\}$. The details of skill library will be introduced at the following subsection. 
\end{itemize}

Based on the above definitions, the problem that the proposed XUAT-Copilot aims to solve is: given a raw test case $C$ associated with an associated parameter list $P$, an interactive test environment $E$ and a predefined skill library $F$, the goal of the proposed system is generate a sequence of actions $A=\left[A_1, ..., A_n\right]$ by continuously interacting with the environment $E$. Since there may be a series of actions for each step of the test case, the generated $A_i$ at $i$th step is still a sequence of actions $A_i = \left[a^1_i ..., a^l_i \right] = \left[f^1_i(p_i^1), ..., f^l_i(p_i^l) \right]$, where each operation $a_i^j$ can be specified as a call of skill function $f_i^j \in F$ associated with a required parameter $p_i^j \in P$.

\subsection{Perception Module}
\begin{table*}[htbp]
  \begin{tabular}{m{1.5cm}|m{9cm}|m{6.5cm}}
  \hline
     & \makecell[c]{\textbf{Raw Contents}} & \makecell[c]{\textbf{Rewritten Contents}} \\ \hline
    \textbf{Template} & System requests User to [PHRASE1], and System validates the result feedback from User is [PHRASE2].  & [PHRASE2'] \\ \hline
    \textbf{Example 1} & System requests User to select a bank for card-less binding, and System validates the result feedback from User is selecting bank. &  Select \$\{bank\_name\}\\ \hline
    \textbf{Example 2} & System requests User to set payment password, and System validates the result feedback from User is submitting payment password. &  Submit payment password (Use numeric keyboard)\\ \hline
  \end{tabular}
  \caption{Templates and Examples of Test Case Before and After Rewriting}
  \label{tab:rewrite}
\end{table*}
Similar to the use of sensors in the field of robotics to observe the state of the agents in the environment, we define a perception module to collect and represent the testing state interactively. In this scenario, the state of the agents in the environment is the GUI page where the current step is located in the APP. Therefore, the function of the perception module is to collect information from the current GUI page and represent it in a certain form. Generally, there are two ways to represent App GUI:
\begin{itemize}[leftmargin=0.5cm]
\item GUI View Hierarchy: A view hierarchy file presenting objects of current GUI page in XML format.
\item GUI Image: A screenshot image of current GUI page.
\end{itemize}
In this paper, we use information of both view hierarchy and screenshot image for state perception. 

The structured text representation of view hierarchy naturally suits LLM Agents. Models such as GPT-3.5 and GPT-4 can understand view hierarchy very well in the setting of in-context learning. The view hierarchy file includes the widget information (coordinates, ID, widget type, text description, etc.), and the layout information on the current GUI page. The state extractor could easily obtain the view hierarchy file of current GUI page through “uiautomator dump” of the Android Debug Bridge (ADB) command [2]. However, the raw view hierarchy has too many tokens easily reaching to the limitation of LLMs. Additionally, redundant information can interfere with LLM's decision-making and reduce accuracy. We define filtering rules in the state extractor to extract concise and informative representation. For each widget node, six attributes, i.e., \texttt{class}, \texttt{resource-id}, \texttt{content-desc}, \texttt{text}, \texttt{clickable}, \texttt{scrollable}, \texttt{bounds} are extracted. \texttt{class} denotes the widget type, \texttt{resource-id} is the ID of the widget, \texttt{content-desc} and \texttt{text} are text description of the widget, and \texttt{bounds} represents the coordinates of the widget. Only widget nodes with non-empty \texttt{text} or \texttt{content-desc} attribute are kept. Meanwhile, \texttt{resource-id} is mocked if the original one is empty since it is very important for LLMs to locate the widget. After filtering, the state is represented as a set of widget node information: $s=\{w_1, ..., w_k\}$, where a widget node $w_i$ is a string with following structure: \\
"\texttt{<node class=, resource-id=, clickable=, scrollable= , content-desc=, text=, bounds= node/>}".

There is a special kind of clickable widgets containing a piece of hyperlinked text and valid operation for these widgets is clicking the hyperlinked text. For example, in Fig.\ref{fig:hyperlink}, there are two widgets with hyperlinked text and their original bounds defined in the view hierarchy are two rectangle areas containing the whole paragraphs with the hyperlinked text. Given this information, LLMs would infer to click the center point based on common sense, but the operation would not be executed successfully since the hyperlinked text are not located at the center of the widget. Therefore, in addition to view hierarchy, we make use of screen images of GUI pages to extract correct clickable bounds of these special widgets. The idea is illustrated as Fig.\ref{fig:hyperlink}. Hyperlinked text always has a unified and unique color different from that of normal text. Based on this prior knowledge, a preprocessing is designed to extract coarse-grained regions of the hyperlinked text. The raw image is firstly filtered and transferred from RGB color space to HSV color space that allows for easy manipulation of hue, saturation, and value, making it suitable for color selection in various applications. Afterwards, the image is binarized by computing the differences between HSV values the image and those of target color and then dilated to get rectangle-like bright regions of the hyperlinked text. Subsequently, rectangle contours are detected in the processed image and used to mask areas without hyperlink in the raw image. Given the masked image, a detection model and a recognition model are used to find the exact bounds, i.e., regions of interest (ROIs), of the hyperlinked text, and recognize the text in each ROI, respectively. As for detection, a pretrained SegLink++ \cite{tang2019seglink++} model is employed. SegLink++ adopts a bottom-up approach, first detecting the attraction-repulsion relationships between text blocks and text lines, then clustering the text blocks into lines, and finally outputting the coordinates of the bounding boxes of the text lines. Based on the detected bounding boxes, a ConvNeXts \cite{liu2022convnet} model is adopted for text recognition. The architecture of ConvNeXts consists of a convolutional backbone and a series of ConvTransformer Blocks. If a widget has more than one detected bounding boxes after image processing, the coordinate of the bounding box with the largest width are used as the new "\texttt{bounds}" attribute of the widget. The recognized text is also used to update the "\texttt{content-desc}" and "\texttt{text}" attributes of the widget.

\subsection{Rewriting Module}
As defined above, a raw test case is an ordered sequence of instructions $C = \left[c_1, ..., c_n \right]$. As shown in Table \ref{tab:rewrite}, each instruction $c_i$ is a piece of text with a fixed template. This kind of expression limits "natural" degree of the instructions. According to the above example, both [PHRASE1] and [PHRASE2] seem to describe the expected users' actions, this may make LLMs confused. In essence, [PHRASE1] in the template is designed for expressing the "process", while [PHRASE2] is supposed to emphasis the "result" of the actions. It is observed that [PHRASE1] often has a very similar description to the title of the current UI page. In other words, [PHRASE1] can be regarded as context information while [PHRASE2] can be regarded as actual user actions expected by the test case. Therefore, rewriting module extracts reformulate the instruction with the following process:
\begin{itemize}[leftmargin=0.5cm]
\item Extracting [PHRASE1] and [PHRASE2] based on regular expression matching.
\item Rewriting [PHRASE2] as [PHRASE2'] with a series of rules based on the information of original [PHRASE1] and [PHRASE2].
\item Use the rewritten [PHRASE2'] as the input instruction for the agents.
\end{itemize}

Due to the page limitation, the details of rewriting rules would not be discussed in this paper. Two typical examples of rewriting are given in Table \ref{tab:rewrite}. The first example involves a specific parameter, thus the rewriting uses a special mark to label the expression of the parameter. For the second example, a rule of information supplementary is utilized. In the WeChat Pay APP, the numeric keyboard pop-up on the screen is the only acceptable way for inputting the payment password. But LLMs may not know this piece of specific knowledge, thus the rewriting rule makes it more clear avoiding LLMs responds with an illegal command.

\subsection{Multi-Agent Collaborative Framework}
The motivation behind the multi-agent framework is the limited capability of a single LLM-based agent on this task. First, inputs of the task, e.g., filtered hierarchy file or parameter list, are long text, which occupy the working memory of the single agent in a large extent. This may easily lead the agent to forgetting, hallucinating and inefficient planning due to the token limitation of LLMs. Additionally, compared with other random-based black-box testing, UAT on WeChat Pay requires for more complex and accurate actions, e.g., selecting a given bank card or inputting a given ID number at correct input widget. Therefore, in order to reduce the complexity of the task and the length of the inputs given to the agents, we decompose the task as three sub-tasks, i.e., action generation, parameter selection and state inspection, and carefully devise the corresponding agents. Specifically, \textbf{Operation Agent} is responsible for planning expected user actions and generating corresponding action commands. \textbf{Parameter Selection Agent} has obligation of choosing appropriate parameters from hundreds of candidates. \textbf{Inspection Agent} is proposed to verify whether the state achieves the goal of current step. 

\begin{algorithm}[t]
\caption{Multi-Agent Instruct2Act}\label{alg:learning}
\Input{Instruction sequence of a test case $C=\left[c_1, ..., c_n\right]$, parameter list $P$, language descriptions $D_{F}$ of skill library $F$, test environment $E$, a backbone large language model $\pi$.}
\Output {Action sequence $A=\left[f_1^1(p_1^1), ..., f_i^t(p_i^t), ...  \right]$.}
Initialize action sequence: $A \gets \left[\right]$ \;
\For{$i = 1,...,n$}{
    Initialize action count within step: $t \gets 1$\; 
    Initialize step goal accomplishment flag: $g \gets 0$\;
    Initialize system memory: $M \gets \emptyset$ \;
    Initialize action sequence: $A_i \gets \left[\right]$ \;
    Observe state $s_i^t$ from environment $E$ \;
    \While{$g \neq 1$}{
        $f_i^t \gets \pi(M, D_{F}, c_i, s_i^t, prompt_{op})$,  $f_i^t \in F$\;
        $p_i^t \gets \pi(P, c_i, f_i^t, prompt_{para})$,  $p_i^t \in P$\;
        Execute action $f_i^t(p_i^t)$ in environment $E$\;
        Initialize invalid history: $H \gets \left[\right]$ \;
        \While{$f_i^t(p_i^t)$ \textup{is invalid}}{
            $H \gets H.\textup{append}(f_i^t(p_i^t))$\;
            $prompt^\prime_{op} \gets prompt_{op} \cup prompt_{relf} \cup H$\;
            $f_i^t \gets \pi(M, D_{F}, c_i, s_i^t, prompt^\prime_{op})$,  $f_i^t \in F$\;
            $p_i^t \gets \pi(P, c_i, f_i^t, prompt_{para})$,  $p_i^t \in P$\;
            Execute action $f_i^t(p_i^t)$ in environment $E$\;
        }
        $A_i \gets A_i.\textup{append}(f_i^t(p_i^t))$\;
        $t \gets t + 1$\;
        Observe state $s_i^t$ from environment $E$ \;
        $g \gets \pi(A_i, c_{i+1}, s_i^t, prompt_{insp})$\;
        $M \gets \pi(M, f_i^t(p_i^t), s_i^t, g, prompt_{sum})$\;
    }
    $A \gets A.\textup{append}(A_i)$\;
}
\Return $A$
\end{algorithm}

\begin{table*}[htbp]
  \begin{tabular}{|m{6.5cm}|m{11cm}|}
  \hline
   \makecell[c]{\textbf{Skill Functions}}  & \makecell[c]{\textbf{Descriptions}} \\ \hline
    \makecell[l]{\texttt{\textbf{click}(rid: str)}}  & Click a UI element specified by the resource id (rid) of view hierarchy \\ \hline
    \makecell[l]{\texttt{\textbf{input\_text}(rid: str, text: str)}}  & Input the content string into a text box specified by the resource id of view hierarchy \\ \hline
    \makecell[l]{\texttt{\textbf{input\_by\_numeric\_keyboard}(digits: str)}}  & Use the numeric keyboard to enter the digits \\ \hline
    \makecell[l]{\texttt{\textbf{press\_adb\_back\_key}()}}  & Press back key on Android device, this is different from clicking back button on screen \\ \hline
    \makecell[l]{\texttt{\textbf{scroll}(direction: str)}}  & Scroll on screen \\ \hline
    \makecell[l]{\texttt{\textbf{swipe\_selector}(rid: str, direction: str)}}  & Swipe a selector specified by specified by the resource id of view hierarchy \\ \hline
  \end{tabular}
  \caption{Skill Library}
  \label{tab:skill}
\end{table*}
\subsubsection{Multi-Agent Instruct2Act Algorithm}

The detailed collaborative mechanism of the proposed multi-agent system is summarized as Alg.\ref{alg:learning}. Apart from the above-mentioned definitions, the backbone LLM that the three agents are founded on is denoted as $\pi$. At the beginning of each step, the algorithm initializes the system memory $M$ and the corresponding action sequence of this step $A_i$ as empty and sets the step goal as unaccomplished state. Afterwards, the algorithm continuously generates actions for the current step until the goal is accomplished. At each iteration of action generation, the operation agent, which is designed by prompting the backbone LLM $\pi$ with $prompt_{op}$, firstly generates a skill(operation) function $f_i^t$ given the memory $M$, the descriptions of the skill library $D_F$, the instruction of the current step $c_i$, and the state $s_i^t$. Based on $c_i$ and the generated $f_i^t$, the parameter selection agent is activated by prompting $\pi$ with $prompt_{para}$ and select an appropriate parameter for $f_i^t$ from the parameter list $P$. A complete action is the inferred operation function associated with the selected parameter, i.e., $f_i^t(p_i^t)$. However, actions generated by LLM may not always be valid in the test environment, e.g., non-predefined functions. Therefore, a self-reflection mechanism is introduced to make the operation agent correct itself based on the reflection prompt $prompt_{relf}$ and the invalid history until a valid action $f_i^t(p_i^t)$ is generated. Afterwards, the action sequence of the current step $A_i$ is appended with $f_i^t(p_i^t)$, and a new state $s_i^t$ is captured. Given the updated $A_i$, $s_i^t$ and the instruction of next step $c_{i+1}$, the inspection agent makes a judgement on whether the step goal is accomplished by prompting $\pi$ with $prompt_{insp}$. Subsequently, the operation agent updates the memory $M$ by summarizing the conversation with $prompt_{sum}$. If the step goal is accomplished, the algorithm will start a new iteration for action generation of next step. It is worth mentioning that, in spite of using different prompts, summarization and self-reflection are still the behavior of the operation agent, since they continuously maintain and use the common memory $M$ in one conversational session. Therefore, the operation agent is the core controller of the system. After goals of all steps are accomplished, the algorithm will return a full sequence of all recorded actions as the final output.

\subsubsection{Skill Library}
A skill is defined as a high-level operation function wrapping ADB commands to simulate users’ actions on the APP. Natural language descriptions are carefully prompted to make these skills more understandable for the agent. Skill functions and their descriptions are defined as Table \ref{tab:skill}. As for \texttt{\textbf{click}}, \texttt{\textbf{input\_text}}, \texttt{\textbf{swipe\_selector}}, the operation agent needs to identify the resource id of the target widget, which is an argument for these skill functions. When using input functions, i.e., \texttt{\textbf{input\_text}} and \texttt{\textbf{input\_by\_numeric\_keyboard}}, the operation agent asks the parameter selection agent for test parameters as inputs of the two functions. We use the following template to embed the information of skill library into the prompts of the operation agent:\\
"\texttt{Exclusively use commands listed below, e.g. command\_name. Commands:}\\
\texttt{[NAME]: [DESCRIPTION], args: ([ARGUMENT]: [TYPE], ...);}\\
...... \\
\texttt{[NAME]: [DESCRIPTION], args: ([ARGUMENT]: [TYPE], ...).}"\\

\subsubsection{Prompt Structure}
We use a shared prompt structure for consistency, yet with agent-specific content. The prompt is composed of the following key components:
\begin{itemize}
    \item \textbf{Profiling}: indicates a specific role of the agent.
    \item \textbf{Task}: describes detailed task of the agent including requirements and constraints.
    \item \textbf{Capability}: represents the agent’s available skills or prior knowledge. 
    \item \textbf{Instruction}: is rewritten instruction of a specific test case step. 
    \item \textbf{Observation}: denotes observed state (GUI information) in the environment. 
    \item \textbf{Context}: includes all useful context information, e.g., past dialog or information from other agents.
    \item \textbf{Response Format}: defines a specific JSON template of the agent response, which has the following elements:
    \begin{itemize}
        \item \textbf{Reasoning}: asks for a detailed reasoning process of the inferred answer.
        \item \textbf{Plan}: (optional) requires for detailed action plan, which is only used for the operation agent.
        \item \textbf{Answer}: specifies the structure and elements of the answer. This varies among agents.
    \end{itemize}
\end{itemize}

For each agent, profiling, task, capability and response format are fixed all the time, while instruction, observation and context are dynamically collected.

\subsubsection{Operation Agent}
As the core of the system, operation agent is responsible for planning expected user actions and generating corresponding action functions. The contents of ``Capability" component are skill functions and corresponding descriptions. ``Instruction" is the rewritten text of current step, and ``Observation" is the text representation of current GUI page. ``Context" component of operation agent is relatively complex, including:
\begin{itemize}
    \item \textbf{Working Memory}: past dialog and executed actions from previous rounds.
    \item \textbf{Invalid Actions}: (optional) invalid actions and corresponding results at the current step. 
    \item \textbf{Reflection}: (optional) prompts the agent to review invalid actions and provide self reflection.
\end{itemize}
The expected response is an action through a calling of a specific skill function, thus ``Answer" of operation agent requires for skill function name, corresponding arguments and natural language description of this action.

\subsubsection{Parameter Selection Agent}
Parameter selection agent is responsible for choosing appropriate parameters from hundreds of candidates. ``Capability" of parameter selection agent is the parameter list. Since parameter selection agent only need to interact with operation agent, ``Observation" is empty and ``Context" is information of inferred skill function from operation agent. Only when inferred skills are inputting functions, parameter selection agent is activated. ``Answer" of parameter selection agent specifies the structure should be the name of the selected parameter.

\subsubsection{Inspection Agent}
Inspection agent aims at verifying whether the state achieves the goal of current step. Different from operation agent, ``Instruction" and ``Observation" of inspection agent are the content of next step and the information of GUI page after actions, respectively. ``Context" of inspection agent is the sequence of performed actions of previous steps, Since the expected answer of inspection agent is binary, i.e., goal-accomplished or not, ``Answer" of inspection agent is either "yes" or "no".

\section{Experiments}
\subsection{Evaluation Settings}
\begin{table}[htbp]
  \begin{tabular}{c|c|c}
  \hline
   \textbf{Number of Test Cases}  & \textbf{Average Steps} & \textbf{Average Actions} \\ \hline
   450  & 7 & 15 \\ \hline
  \end{tabular}
  \caption{Statistics of Emperimental Data}
  \label{tab:data}
\end{table}

\paragraph{\textbf{Experimental Data}}
The objective of the proposed model is to generate correct scripts of test cases in the UAT system. Test cases without bugs or other system problems are sampled from the system as evaluation data. This means that all the experimental cases have passed the UAT with human implemented scripts. The statistics of the experimental data is summarized in Table \ref{tab:data}.
\paragraph{\textbf{Metrics}}
The main evaluation metrics are $Pass@1$ and $Complete@1$. $@1$ denotes allowing the method to generate candidate script for each test case only once. $Pass@1$ represents case-level pass rate, i.e., the ratio of successfully passed cases among all test cases: 
\[
    Pass@1 = \frac{\sum_{k=1}^N\mathbbm{1}[C^k \textup{ is passed}]}{N}
\]
where $C^k$ is $k$th test case in the experimental dataset. $\mathbbm{1}(C^k)$ is an indicator function returning 1 if the whole test case $C^k$ is passed and 0 otherwise.

$Complete@1$ represents step-level complete rate, i.e., the average ratio of successfully completed steps on all test cases: 
\[
    Complete@1 = \frac{\sum_{k=1}^N\sum_{j=1}^{L_k}\mathbbm{1}[c_j^k \textup{ is passed}]}{\sum_{k=1}^N L_k}
\]
where $N$ is the total number of test cases, $L_k$ represents the length (number of steps) of $k$th test case and $c_j^k$ denotes $j$th step of $k$th test case. Similarly, $\mathbbm{1}(c_j^k)$ is 1 if $c_j^k$ is passed and 0 otherwise.

\paragraph{\textbf{Compared Methods}}
In the literature, there is few research works or public methods for automated UAT, particularly for the stage of test script generation. Therefore, in this paper, we compare the proposed method with its variants.
\begin{itemize}[leftmargin=0.5cm]
    \item \textbf{Single Agent}: this is a variant that only one agent is used and it should fulfill all sub-tasks when generation actions. The prompt consists of all agents' prompts from XUAT-Copilot.
    \item \textbf{Without Reflection}: this is a variant that reflection part is removed from the prompt of operation agent. 
\end{itemize}
\subsection{Evaluation Performance}
\begin{table}[htbp]
  \begin{tabular}{c|c|c}
  \hline
   \textbf{Methods}  & $Pass@1$ & $Complete@1$ \\ \hline
   Single Agent  & 22.65\% & 25.25\% \\ \hline
   Without Reflection  & 81.96\% & 89.39\% \\ \hline
   \textbf{XUAT-Copilot} & \textbf{88.55\%}  & \textbf{93.03\%} \\ \hline
  \end{tabular}
  \caption{Evaluation Performance}
  \label{tab:eval}
\end{table}
The evaluation performance is listed in Table \ref{tab:eval}. Compared with the single agent version, XUAT-Copilot gains significant improvements on both metrics. This demonstrates the superiority of the proposed multi-agent framework based on LLM. As mentioned above, single agent cannot handle multiple tasks in the UAT processes simultaneously, and long prompts may easily lead to forgetting, hallucinating and inefficient planning. Additionally, the experimental results prove the importance of reflections in prompts. When the reflection is removed, the evaluation performance decreases.

\section{Conclusion}
In this paper, we investigate the possibility and feasibility of LLMs to automate the test script generation in the UAT process. A multi-agent system XUAT-Copilot is proposed for this purpose. Three agents in the proposed system handle different tasks for test script generation and collaborate with each other. The superiority of the proposed system is demonstrated by the experiments. The system has launched the formal testing environment of WeChat Pay APP to improve the efficiency of test script generation. As for future work, we plan to further improve the generalization ability of the system by removing some rule-based parts and introducing more LLM agents for more specific tasks. 

\bibliographystyle{ACM-Reference-Format}
\bibliography{sample-base}

\end{document}